\documentclass{article}

\usepackage[preprint]{neurips_2024}

\usepackage[utf8]{inputenc} 
\usepackage[T1]{fontenc}    
\usepackage{hyperref}       
\usepackage{url}            
\usepackage{booktabs}       
\usepackage{amsfonts}       
\usepackage{nicefrac}       
\usepackage{microtype}      
\usepackage{xcolor}         
\usepackage{threeparttable}
\usepackage{ulem}
\usepackage{amsmath}
\usepackage{algorithm}
\usepackage{algorithmic}
\usepackage{mdframed}
\usepackage{amssymb}
\usepackage{multirow}
\usepackage{tabularx}
\usepackage{booktabs}
\usepackage{geometry}
\usepackage{longtable}
\usepackage{times,a4wide}
\usepackage{graphicx}
\usepackage{amsmath}
\usepackage{listings}
\usepackage{tabularx}
\usepackage[most]{tcolorbox}
\usepackage{subfigure}
\usepackage[capitalise]{cleveref}
\usepackage{xspace}
\usepackage{enumitem}

\newcommand{\attentionname}{Round Attention\xspace}

\usepackage{caption}
\usepackage{listings}
\usepackage{float}

\title{Round Attention: A Novel Round-Level Attention Mechanism to Accelerate LLM Inference}

\newcommand*\samethanks[1][\value{footnote}]{\footnotemark[#1]}
\author{
	Yaohua Tang\thanks{Equal contribution. tangyaohua28@gmail.com}\quad  
	Zhicheng Hu\samethanks{} \quad
	Kun Cheng\samethanks{}
	 \\
	\textbf{
		Fan Mo \quad
        Qiheng Lv   \quad
		Hua Wang \thanks{Corresponding author. \texttt{wangtianyu.di@gmail.com}}\quad
        Zhi Chen \thanks{\texttt{zhic@mthreads.com}}
        } 
	\\\\
	Moore Threads AI
	\\
	 }

\begin{document}

\maketitle

\begin{abstract}
The increasing context window size in large language models (LLMs) has improved their ability to handle complex, long-text tasks.  However,  as the conversation rounds continue, it is required to store a large amount of KV cache in GPU memory, which significantly affects the efficiency and even availability of the model serving systems. This paper analyzes dialogue data from real users on the granularity of round and discovers that the LLM inference manifests a watershed layer, after which the distribution of round-level attention shows notable similarity. 
Based on this, we propose Round Attention - a novel round-level attention mechanism that selectively processes the KV cache of top-k relevant rounds, where k is dynamically determined through the attention matrix in the watershed layer. Theoretical analysis demonstrates that our method reduces memory usage by 54\% to 82\%, while experimental results confirm that loading sparse critical-round KV cache maintains answer accuracy without performance degradation.
\end{abstract}

	\section{Introduction}
	
	\label{introduction}

Recent advancements in large language models have facilitated the wider adoption of language model services for everyday problem-solving tasks. However, prolonged interactions expose two significant challenges. First, the rapid expansion of context length incurs substantial computational overhead due to the quadratic scaling of self-attention mechanisms. Second, although key-value (KV) caching alleviates redundant computations, it substantially increases GPU memory requirements, resulting in limited inference batch sizes and GPU under-utilization. For instance, an NVIDIA A100 with 40GB of memory can accommodate only a single LLaMA request with a context length of 128K, spending nearly $50\%$ of its processing time on KV cache access \citep{he2024fastdecode}.

\begin{figure*}
    \centering
    \includegraphics[width=0.9\linewidth]{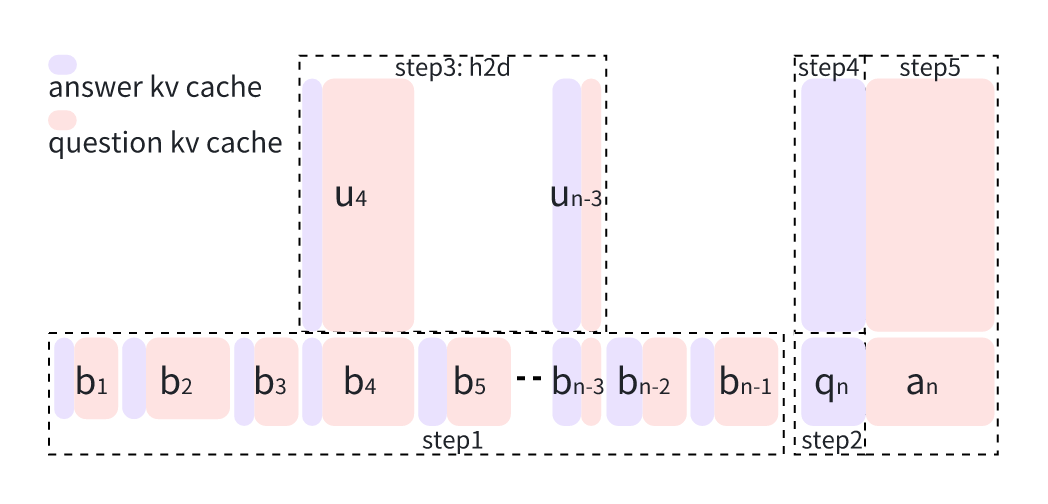}
    \caption{The inference pipeline of Round Attention. Our KV cache is managed and stored in a round-based manner. For a given token, the KV cache is divided into two tensors: the upper and lower halves. The complete KV cache is offloaded to CPU memory, while only the lower half tensor is retained in GPU memory. The upper half tensor is transferred from CPU memory to GPU memory at real-time based on query relevance, thereby optimizing memory usage.}
    \label{fig:pipeline}
\end{figure*}

To enhance inference efficiency,  previous research has investigated KV cache eviction and sparse attention techniques for LLMs,  noting that attention is inherently sparse. These methods either store the entire KV cache in GPU memory,  selecting key tokens during auto-regression to reduce cross-attention computation time \citep{tang2024quest},  or maintain the KV cache in CPU memory, transferring it to GPU memory token by token during inference \citep{chen2024magicpig, sun2024shadowkv,  he2024fastdecode,  lee2024infinigenefficientgenerativeinference}. The former does not reduce GPU memory usage,  while the latter incurs significant communication overhead. Furthermore, current methods often require an expensive calculation of the most relevant tokens for each layer.

Another common issue with the aforementioned studies is that they restrict their analysis of contextual relationships to the token level. Analysis in \cite{sun2024shadowkv} reveals that most post-RoPE keys exhibit high cosine similarity with adjacent tokens, enabling chunk-level approximations for selecting important tokens.  LONGMEMEVAL benchmark\citep{wu2024longmemevalbenchmarkingchatassistants} explores the memory design options for memory-augmented chat assistants and discovers that round is ``the best'' granularity for storing and utilizing the interactive history. Inspired by this, we analyze the attention matrix under round granularity and identify two interesting patterns. First, the attention score distributions at the round granularity in prevalent open-source large models exhibit considerable variability in the initial layers; however, from a certain layer onward, the distributions between layers become remarkably similar. Second, within a single dialogue round, the attention scores computed for the ``question'' in relation to previous dialogue turns closely resemble those computed for the ``answer'' in relation to previous dialogue turns. 

Building on these observations, we propose \textbf{\attentionname}, a method that leverages the sparsity of the attention matrix. Specifically, in Round Attention, the KV caches for only the initial layers are retained in GPU memory, while those for the deeper layers are offloaded to CPU memory. During inference, for each question in a dialogue round, we compute the attention scores between the current question and previous dialogue rounds, and then collectively load the KV caches of the top-k rounds from CPU memory back into GPU memory to facilitate subsequent computations. This strategy enables us to substantially reduce GPU memory consumption.  Due to the first identified pattern, we only need to compute the top-k rounds once at a specific layer and then perform a single host-to-device (h2d) operation to transfer the corresponding KV cache tensor to GPU memory. This approach contrasts with other methods that require top-k computations at each layer and transfer the KV cache at the token granularity, significantly reducing the latency overhead associated with top-k calculations and offloading mentioned in other approaches.

Our primary contributions are as follows:
\begin{itemize}
    \item We dissect the attention patterns in LLM post-deployment at the round  granularity and reveal two enlightening characteristics in attention matrix in real applications.
    \item Based on these characteristics,  we design a novel method, Round Attention, associated with an array of techniques for long-context dialogues. This approach stores and transfers the KV cache at round granularity.
    \item We conduct extensive experiments on the proposed approach. The results show that it can reduce the GPU memory footprint by $54\%$ to $82\%$ with no accuracy loss. More importantly, thanks to the one-time top-k selection and host-to-device (h2d) transfer, our method achieves lower latency compared to standard non-offloaded Flash Attention.
\end{itemize}

\section{Related Work}
\subsection{Attention Matrix Analysis}
The sparsity of attention weights in pre-trained LLMs,  especially in long-context scenarios,  has been well-documented \citep{liu2022dynamic,  ribar2023sparq, liu2023deja,  xiao2023efficient}. 
\citet{ma2024compressing} investigates the distribution of important tokens in the context and discovered that recent tokens are more important than distant ones. They also find that attention scores between consecutive layers are similar, which has also been previously observed in smaller models \citep{xiao2019sharing,  bhojanapalli2021leveraging}.

\citet{mu2024cross} reports that Attention weights were remarkably similar between the transformer layers,  particularly the adjacent layers. \citet{men2024shortgpt} identify notable redundancy across LLM layers,  where some layers contribute marginally to the model.  \citet{fan2024not} shows that for some tasks,  LLMs can achieve results comparable to the final output at some intermediate layers.

\subsection{KV Cache Eviction Algorithm}

Many previous efforts focuse on KV cache compression to accelerate attention and reduce memory usage. H2O \citep{zhang2023h2o} retains a limited budget for the important KV cache regarding the sum of historical attention scores. FastGen \citep{ge2023model} further categorizes tokens and only keeps partial KV cache using a more sophisticated strategy. TOVA \citep{oren2024transformers} simplifies the policy by determining the permanently discarded tokens using the current query. StreamingLLM \citep{xiao2023efficient} handles infinitely long text with attention sinks and a finite KV cache. SparQ \citep{ribar2023sparq} computes approximate attention scores by channel pruning and selects important tokens through them. \citep{tang2024quest} concludes that the importance of a token is highly dependent on the query and proposes Quest, a method that records the \textit{min} and \textit{max} key values in KV cache pages and estimates the importance of a page using query vectors.

However, these approaches face several challenges. First, it is costly to identify the top-k attention. For example, applying a naive search algorithm, e.g. IVF \citep{douze2024faisslibrary}, requires access over $30\%$ key states to obtain the top-k results \citep{liu2024retrievalattention}, which is quite compute-intensive. Second, these approaches save the KV cache in the GPU memory to avoid loading them from the CPU memory, which does not reduce the total memory consumption of KV cache, hence limiting the max context window and inference batch size.

Some papers attempted to offload KV cache to CPU memory to reduce the active GPU memory usage. \cite{liu2024retrievalattention} proposes to build approximate nearest neighbor search (ANNS) indexes for KV vectors in CPU memory and retrieve the most relevant ones through vector search during generation. \cite{sun2024shadowkv} stores the low-rank key cache and offloads the value cache to reduce the memory footprint for larger batch sizes and longer sequences. \cite{chen2024magicpig} stores the LSH hash tables and runs the attention computation on the CPU, which significantly reduces the workload of attention computation. However, these works transmit the key-value (KV) cache at the token level, and in some approaches, the top-k selection is computed on a per-layer basis, which implies that the KV cache is also transferred layer by layer, resulting in significant overhead for h2d transfers.

\section{Methodology}
This section presents Round Attention,  a novel approach that dissects the attention matrix at the round level for multi-round dialogue tasks by taking $<q, a>$ pairs as the basic analysis unit. The objective is to reduce the memory footprint and inference latency without sacrificing the accuracy of LLMs.

\subsection{Attention Distribution}
\label{sec:attentiondist}
Given an input sequence $X = [x_1, x_2, \dots]$,  a standard Transformer \citep{vaswani2023attentionneed} network computes a set of queries $Q$,  keys $K$,  and values $V$ using linear transformations on $X$. It then computes the self-attention scores as $\text{Att}(Q, K) = \text{softmax}(\frac{QK^T}{\sqrt{d_k}})$. To investigate the attention pattern among rounds, we denote the sum of the attention scores of the tokens in $\textbf{q}_n$, $\textbf{a}_n$ and the tokens in $<\textbf{q}_k,  \textbf{a}_k>$ of the previous $k$-th round for layer $l$ as:

\begin{equation}
\setlength\abovedisplayskip{3pt}
\setlength\belowdisplayskip{3pt}
\label{equ:qattention}
  \text{qAtt}_k^l = \sum_{\substack{i \in \textbf{q}_n,j \in <\textbf{q}_k,  \textbf{a}_k>}}\text{Att}(Q_i^l, K_j^l), \quad \text{aAtt}_k^l = \sum_{\substack{i \in \textbf{a}_n, j \in <\textbf{q}_k,  \textbf{a}_k>}}\text{Att}(Q_i^l, K_j^l)
\end{equation}


The distribution $P_q^l$ is calculated by normalizing $\text{qAtt}_k^l$, and $P_a^l$ is calculated by normalizing $\text{aAtt}_k^l$. We examine the distribution patterns of $P_q^l$ and $P_a^l$ within the same layer, as well as the distribution patterns of $P_q^l$ across different layers. SharedGPT\citep{sharegpt}, a dataset produced by conversations between real users and ChatGPT, is adopted to analyze the distribution patterns. Qwen2.5-0.5B \citep{qwen25} is used as the LLM.


\noindent \textbf{Observation 1: Attention distributions of $q_n$ and $a_n$ are similar.}

\begin{figure*}[htbp]
\centering
\subfigure[KL among rounds]{
\includegraphics[width = .475\textwidth]{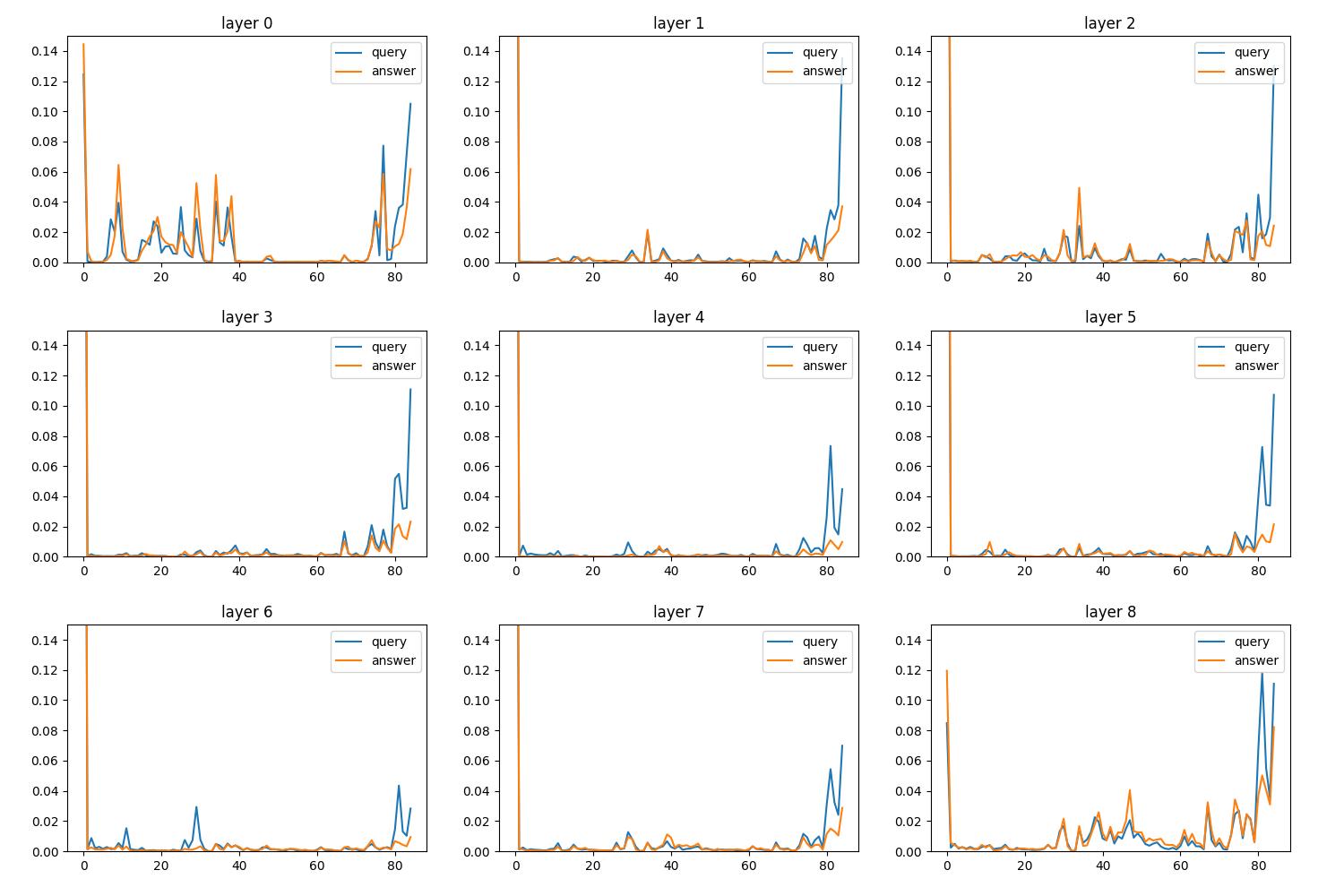}
\label{fig:layerRounds}
}
\subfigure[KL among layers]{
\includegraphics[width = .475\textwidth]{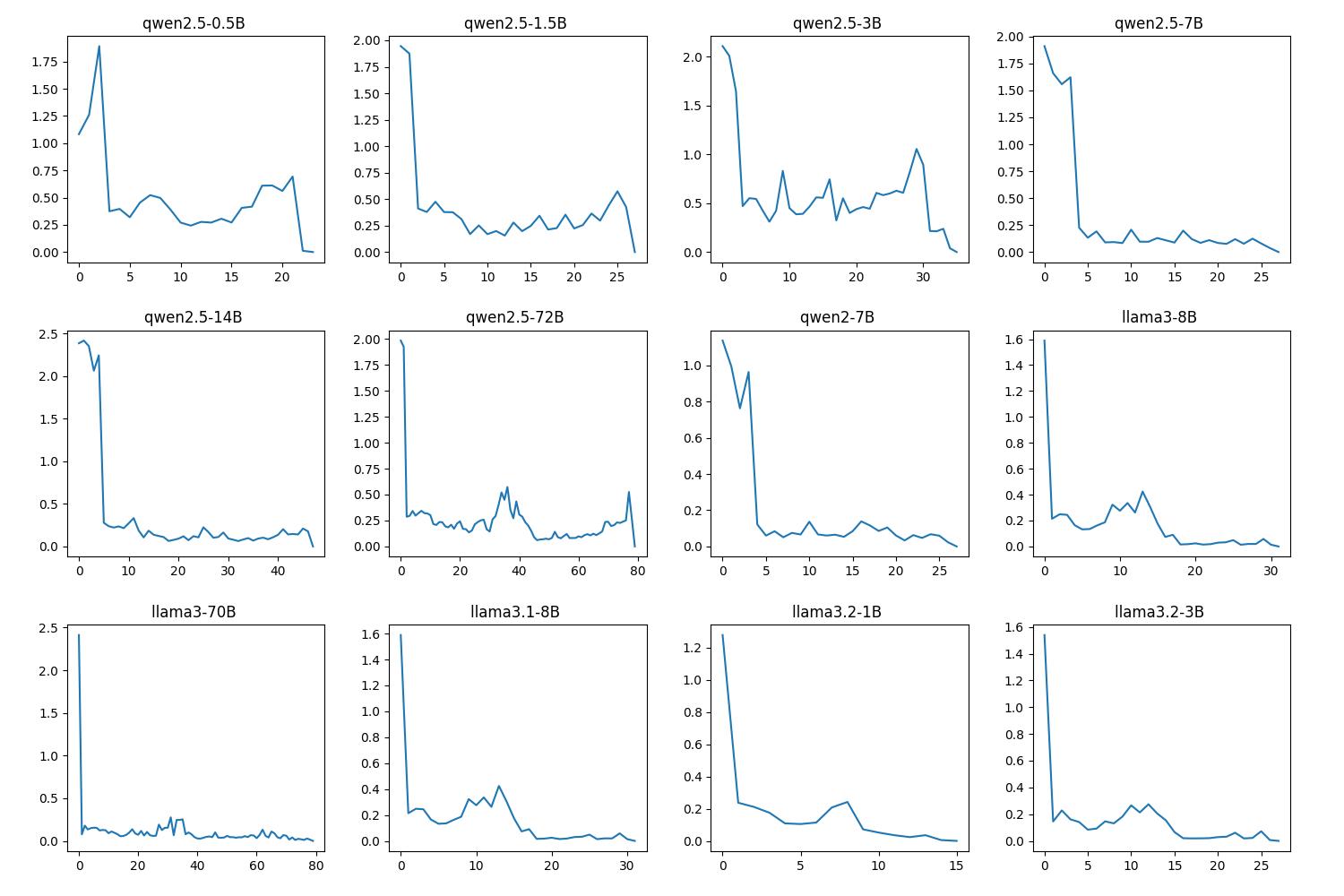}
\label{fig:layerKL}
}
\caption{Round attention distribution patterns. \textbf{(a).} The horizontal axis represents the round index and the vertical axis represents the attention scores of \(q_n/a_n\) in relation to historical rounds. It can be observed that the variation trend of the attention scores calculated for \(q_n\) is highly similar to that of \(a_n\).   \textbf{(b).} The horizontal axis is the layer index. The vertical axis represents the average KL divergence between \(P_q^l\) of each layer and \(P_q^l\) of subsequent layers. It can be observed that nearly all mainstream models exhibit a similar pattern. }
\end{figure*}

As an example, in Figure~\ref{fig:layerRounds}, we selected one dialogue comprising 85 rounds to analyze the attention probability distribution of the 85th round in relation to the preceding rounds across different layers. As shown in the figure, the trends of \(P_q^l\) and \(P_a^l\) are highly similar in each layer, indicating that rounds highly correlated with the question of the 85th round are also highly correlated with its answer. Thus, after performing prefill on the question of the 85th round, we can identify the most relevant historical rounds' KV caches for AR computation based on the round attention distribution, rather than utilizing the KV caches from all rounds. We would like to emphasize that this pattern is not only applicable to this particular example. We have derived such pattern after analyzing a substantial number of dialogues, and we are using this example as a subject for demonstration.

\noindent \textbf{Observation 2: Attention distributions among layers are similar.}
Next, we analyze the correlation of \( P_q^l \) across different layers. For a given layer, we compute the Kullback-Leibler (KL) divergence between that layer and each subsequent layer, averaging these values to obtain the mean KL divergence between the layer and all following layers. We then plot these values for all layers, resulting in Figure~\ref{fig:layerKL}. It can be observed that for nearly all currently mainstream open-source models, regardless of their size, a similar pattern emerges. The initial few layers exhibit significant differences compared to the subsequent layers; however, after reaching a certain layer, the disparity suddenly diminishes substantially. We designate this layer as ``\textbf{watershed layer}'' $L_w$ and we list this layer for several open-source models in Appendix~\ref{app:layer-w}. From this layer onward, the \( P_q^l \) values of the subsequent layers are very close to each other. Although there are occasional instances of slight increases in divergence, these differences remain significantly smaller than those observed in the earlier layers. This indicates that we can select the rounds most relevant to the question at the watershed layer for subsequent attention calculations, thereby eliminating the need to perform this selection computation at every layer, which would incur additional time costs. Based on these two observations, we propose our inference pipeline, Round Attention.

\subsection{Round Attention Inference Pipeline}
\label{sec:pipeline}
Figure~\ref{fig:pipeline} depicts the pipeline for Round Attention. First, we design a strategy to determine the watershed layer for a given LLM. In real multi-turn dialogue LLM serving systems,  it is impractical to store all historical KV caches from all users in the GPU memory. A user's historical KV cache will normally be swapped out to the host memory or even slower storage devices when she is inactive for some period,  so that the precious GPU memory can be well-utilized. For simplicity,  we assume that the LLM has $L$ layers. $b_m$ denotes the KV cache of $1\sim L_w$ layers for the $m$-th dialogue round,  $u_m$ denotes the KV cache of $L_w\sim L$ layers for the $m$-th dialogue round. $b_m$ and $u_m$ are stored as separate tensor in memory. 

When the user becomes active,  e.g. asking LLM the $n$-th question $\textbf{q}_n$,  the following steps will be executed to conduct the inference for this turn.

\begin{itemize}
\setlength{\itemsep}{0pt}
    \item step1: Load $b_1\dots b_{n-1}$ to the GPU memory from the host memory.
    \item step2: Perform prefill computation for $\textbf{q}_n$ on layer $1\sim L_w$.
    \item step3: Select the most relevant top-k dialogue rounds via the strategies proposed in Section~\ref{sec:roundstrategy} with $\text{qAtt}^{L_w}$, load the KV cache for layer $L_{w+1}\sim L$,  $\{u_{m}\}: m \in top-k$.
    \item step4: Finish prefill for the remaining layers.
    \item step5: Decode $\textbf{a}_n$.
\end{itemize}

Compared to the previous works that work on the token level, therefore invoking multiple fragmented KV cache transfers between host and device memory,  our method works at the dialogue round level where a monolithic tensor for all tokens in a prior round is transferred to GPU at once. Upon the accomplishment of the computation of $u_n$ for layer $L_{w+1}\sim L$ in the $n$-th round,  the new KV cache is saved to the host memory as a monolithic tensor as well. Therefore,  our methods reduces the number of expensive host-to-device(H2D) and device-to-host(D2H) data transferring. In addition, moving data in a large chunk is able to better utilize PCIe bandwidth.  The algorithm is summarized in Algorithm~\ref{alg:example}.

\begin{algorithm}[htb]
   \caption{Round Attention}
   \label{alg:example}
\begin{algorithmic}
   \STATE {\bfseries Input:} $b_1\dots b_{n-1}$, $u_1\dots u_{n-1}$ 
   \STATE Initialize: transfer $b_1\dots b_{n-1}$ from host to device memory
   \FOR{$i=n$ {\bfseries to} $\infty$}
   \STATE New query $\textbf{q}_n$,  conduct prefill calculation for layer  $1\sim L_w$
   \STATE Calculate Top-k rounds based on $\text{qAtt}^{L_w}$
   \STATE transfer $\{u_{m}\}: m \in top-k$ to device memory
   \STATE finish prefill and AR calculation
   \STATE transfer $\{u_{m}\}: m \in top-k$ and $u_{n}$ to host memory
   \ENDFOR
   \STATE transfer $b_n\dots b_{\infty}$ to host memory
\end{algorithmic}
\end{algorithm}

\subsection{Round Strategy}
\label{sec:roundstrategy}
Three strategies are considered to determine the top $k$ most relevant dialogue rounds after $L_w$ is discovered and $\text{qAtt}_k^{L_w}$ is computed. 

\noindent\textbf{Strategy 1: Fixed rounds} selects the satisfied rounds using a predefined threshold,  e.g. $\text{qAtt}_k^{L_w}> v$.  Analyzing the distribution of attention scores across rounds we find  that the attention values are concentrated in a limited number of rounds, with the majority of rounds exhibiting minimal attention scores. Thus we selected \( v = 0.1 \).

\noindent\textbf{Strategy 2: top-k rounds} picks that rounds that correspond to the top $10\%$ $\text{qAtt}_k^{L_w}$. Analyzing the distribution of round attention scores reveals that the top $10\%$ of rounds account for over $80\%$ of the cumulative attention.

\noindent\textbf{Strategy 3: Adaptive rounds} chooses the rounds adaptively with the $\text{qAtt}_k^{L_w}$  distribution. The condition is defined as: $\text{qAtt}_k^{L_w}>mean + k*std$,  where $mean$ and $std$ are the mean and standard deviation of $\text{qAtt}^{L_w}$.

\subsection{KV cache Dropping}
\label{sec:cachedropping}
We observed that the KV caches of some dialogue rounds in the ShareGPT are never active and do not affect the inference quality even if removed for attention computation. For these rounds,  we delete the KV cache in the corresponding tokens to avoid saving them.

\subsection{Memory Footprint Analysis}
\label{sec:memcommanalysis}

Given an LLM with context length $S$,  hidden size $H$,  total layers $L$,  and inference batch size $B$,  the amount of memory consumed by the KV cache is calculated by $M_{orig} = 2 * 2 * B * S * H * L$, where the first 2 represents K and V,  and the second 2 means that float16 occupies 2 bytes. For Round Attention,  assuming that the $K$ most relevant rounds are chosen from the total $T$ rounds of dialogue,  the amount of memory used by each round on average is calcualted as $M_{round} = 4B*S*H*L_w + 4B * K/T * S * H * (L-L_w)$. This is because layer $1\sim L_w$ uses the entire KV cache,  and the subsequent layers ($L_{w+1}\sim L$) only compute the attention with the most relevant $K$ rounds. The memory saving ratio of Round Attention can be expressed as Equation~\ref{equ:memoryratio}.

\small
\begin{equation}
\label{equ:memoryratio}
    \frac{M_{round}}{M_{orig}} = \frac{L_w + K*(L-L_w)/T}{L} = 
    \frac{L_w}{L} + \frac{K}{T}(1-\frac{L_w}{L})
\end{equation}
\normalsize
Since $K$ is much smaller than $T$ in practice,  e.g. $6\sim8$ vs tens to hundreds,  the upper bound of Equation~\ref{equ:memoryratio} approximates to $\frac{L_w}{L}$. When $K$ equals $T$,  that is,  all dialogue rounds are selected,  Round Attention degrades to the original inference with virtually no memory cost.
As shown in Table~\ref{app:layer-w}, for mainstream large models, the ratio \(\frac{L_w}{L}\) ranges from 0.18 to 0.46, indicating a memory saving percentage of 54\% to 82\%, which is quite substantial.

\section{Experiments And Analysis}
\subsection{Experiment Setting}
\noindent \textbf{Data.} Two widely used datasets,  ShareGPT\citep{sharegpt} and LONGMEMEVAL\citep{wu2024longmemevalbenchmarkingchatassistants},  are used to evaluate the effectiveness of Round Attention. ShareGPT contains a collection of approximately 52K user-shared conversations scraped through the ShareGPT API. These conversations are multi-turn,  including both user prompts and responses from ChatGPT.  LONGMEMEVAL is a comprehensive benchmark designed to evaluate five core long-term memory capabilities of commercial chat assistants: information extraction,  multi-session reasoning,  temporal reasoning,  knowledge updates,  and abstention. This benchmark also records the historical user-assistant conversations with 250 rounds on average. This is a difficult dataset, on which GPT-4o's accuracy is only $0.5773$.

\noindent \textbf{Baselines.} A suite of the latest open-source LLMs, e.g. Qwen2.5,  LLaMA3,  and LLaMA3.2 \citep{grattafiori2024llama3herdmodels}, are tested on the above datasets, but our approach can be applied to any other long-context LLMs. We use PyTorch and FlashAttention \citep{dao2022flashattentionfastmemoryefficientexact} as the default inference framework, which are refered as \texttt{Flash}. All testing are conducted on a single Nvidia A100 GPU with 80GB of memory, equipped with PCIe. The CPU used was an Intel(R) Xeon(R) Gold 6346 CPU operating at 3.10GHz (1.16/3.60GHz), and the system had 1TB of memory.

\noindent
\begin{minipage}{\textwidth}
\scriptsize
\tabcolsep=0.1cm
\begin{minipage}[t]{0.39\textwidth}
\makeatletter\def\@captype{table}
\caption{Accuracy for Qwen2.5-3B under different round strategies}
\label{table:accuracy}
\begin{tabular}{lccccc}
\toprule
                      &        & \texttt{Flash} & top-k & Fixed & Adaptive \\\midrule
\multirow{2}{*}{\texttt{mini} } & score  &  7.51     &  7.5    & 7.33 & 7.5 \\
                      & tokens &  809     &  515    & 560 & 515\\
\multirow{2}{*}{\texttt{small}} & score  & 7.49      & 7.47     & 7.48 &7.5 \\
                      & tokens &  4339     &  1245    & 1245 &1245 \\
\multirow{2}{*}{\texttt{medium}} & score  & 7.42      & 7.5     & 7.5 & 7.48\\
                      & tokens &  11491     &  1276    & 1089 & 1089\\
\multirow{2}{*}{\texttt{large}} & score  & 7.49      &  7.46    & 7.43 & 7.4\\
                      & tokens &  19548     &  2343    & 1142 & 1639\\
\midrule
Ave & score & 7.477 & \textbf{7.483} & 7.435 & 7.470\\
\bottomrule
\end{tabular}
\end{minipage}
\quad\quad\quad\quad
\begin{minipage}[t]{0.56\textwidth}
\makeatletter\def\@captype{table}
\caption{Accuracy for different Model \\under the top-k round strategy}
\label{table:accuracyformodels}
\begin{tabular}{lcccccc}
\toprule
\multirow{2}{*}{Attribute}   & \multicolumn{2}{c}{\texttt{small}} & \multicolumn{2}{c}{\texttt{medium}} & \multicolumn{2}{c}{\texttt{large}}  \\ 
\cline{2-3} \cline{4-5} \cline{6-7}  
                     & score & tokens & score & tokens & score & tokens \\\midrule
\texttt{Flash}       & \textbf{6.39} & 4339 & \textbf{5.95} & 11491 & 5.8  & 19548 \\ 
Qwen2.5-0.5B   & 5.74 & 1245 & 5.85 & 1199  & \textbf{6.06} & 2391  \\ \midrule
\texttt{Flash}       & \textbf{7.77} & 4339 & \textbf{7.8}  & 11491 & 7.44 & 19548 \\ 
Qwen2.5-7B     & 7.08 & 1218 & 7.49 & 1382  & \textbf{7.57} & 2448  \\ \midrule
\texttt{Flash}       & 7.45 & 4260 & \textbf{4.05} & 11735 & 3.01 & 19812 \\ 
Llama3-8B      & \textbf{7.6}  & 1227 & 3.84 & 1695  & \textbf{3.47} & 3376  \\ \midrule
\texttt{Flash}       & 7.35 & 4260 & 7.11 & 11735 & 7.38 & 19812  \\ 
Llama3.1-8B   & \textbf{7.39} & 1212 & \textbf{7.17} & 1369  & \textbf{7.46} & 2477 \\ 
\bottomrule
\end{tabular}
\end{minipage}
\end{minipage}

\subsection{Accuracy Evaluation}
\label{sec:accu}
We classify ShareGPT into four categories with respect to dialogue rounds,  \texttt{mini} (0-10 rounds),  \texttt{small} (10-30 rounds),  \texttt{medium} (30-50 rounds),  and \texttt{large} (50-100 rounds). We treat the last prompt in each category as $\textbf{q}_n$,  and then use the default inference framework and Round Attention to compute $\textbf{a}_n$. GPT-4o is employed as the Judger to evaluate the quality of the generated results. Each $\textbf{a}_n$ is evaluated 5 times and the average score is taken as the final score of the response. The prompts are chosen from AlignBench \citep{liu2023alignbench},  where the samples can be found in Appendix~\ref{appendix: gpt4judge}.


As shown in Table~\ref{table:accuracy},  the top-k strategy is the most effective,  with an average score exceeding that of the standard inference engine. This indicates that using only the KV caches from the top-k rounds doesn't have notable impact on the quality of model responses,  as the attention matrix is highly sparse,  particularly for the extremely large conversational rounds. Furthermore,  the number of tokens we processed was reduced by $88\%$ compared to the standard inference engine for large rounds,  which suggests a substantial decrease in attention computation and a significant saving in GPU memory.


To validate the generalize ability of this method,  we test the accuracy on various sizes of the Qwen2.5 model,  as well as on some models from Llama3 and Llama3.2. All experiments employed the top-k round strategy. The results are presented in Table~\ref{table:accuracyformodels}. It is evident that the overall scores of the responses generated by our method on these models are comparable to those produced by the standard inference engine. However, we also observe that for the Qwen-2.5 series models, the accuracy of Round Attention may decrease with fewer rounds. In contrast, when the number of rounds exceeds $50$, Round Attention consistently outperforms Flash Attention.

Since the responses from ShareGPT are subjective,  we also utilize the objective dataset, LONGMEMEVAL, as the test bench to further validate the effectiveness of our approach. The original LONGMEMEVAL benchmark evaluate the results yielded by Llama3-8B.  For consistency purpose, we also run tests on the same model. To showcase the generalization of our approach, we conduct the same experiments on Qwen2.5-7B as well.


As shown in Table~\ref{table:accuracyforlongmem}, despite the challenging nature of LONGMEMEVAL, Round Attention performs remarkably well. For the Llama 3-8B model, Round Attention is comparable to Flash Attention, whereas for the Qwen 2.5-7B model, the accuracy of Round Attention is twice that of Flash Attention. Interestingly, in the temporal reasoning tasks, Round Attention consistently outperforms Flash Attention, indicating that excessive information can lead to interference in reasoning tasks. Identifying the key rounds allows for more accurate inference results.

\subsection{GPU Memory Reduction and Latency Reduction}

\begin{figure}[htb]
    \centering
    \includegraphics[width=0.5\columnwidth]{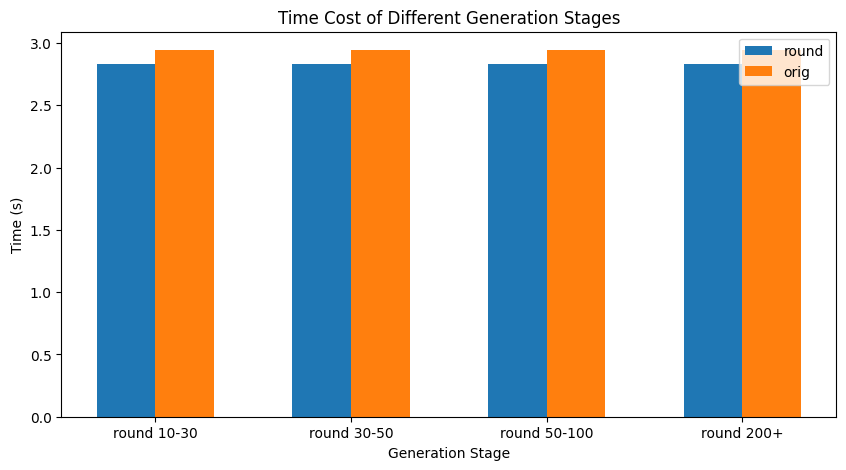}
    \caption{The statistical results of end-to-end inference time for Round Attention compared to Flash Attention across different round categories.}
    \label{fig:latency}
\end{figure}

To empirically evaluate the latency of Round Attention compared to Flash Attention, we selected 20 dialogue samples from each of the four categories mentioned earlier. Each sample was run 10 times, and the average latency was computed and plotted in Figure~\ref{fig:latency}.

It can be observed that for all different round categories, the latency of Round Attention is lower than that of Flash Attention. This improvement is due to our KV cache storage and transfer strategy, which keeps the h2d transfer time manageable, and our top-k selection is computed only once rather than at each layer. We provide a detailed breakdown of latency in the Appendix~\ref{app:latency}. Due to the h2d transfer and the selection of top-k, the latency during the \( q_n \) prefill phase exhibits a slight peak at layer \( L_w \); however, this peak is minor and occurs only once. In contrast, during the \( a_n \) decode phase, the reduction in the KV cache leads to decreased attention computation time. As the number of decode steps increases, this reduction accumulates and ultimately surpasses the one-time overhead from the h2d transfer and top-k selection, resulting in an overall latency that is superior to that of Flash Attention.

\subsection{Round vs Token}

In this section, we compare the two granularities. For both granularities, we employed the same top-k calculation strategy to retrieve the KV cache. Specifically, we first computed the average attention score for all tokens/rounds at each granularity, selecting those with attention scores greater than the average. The model used for testing was Llama3.1-8B. The results are presented in Table~\ref{table:accuracyroundvstokens}. It is evident that, with the same top-k calculation strategy, the recall accuracy at the round granularity surpasses that of the token granularity. This is particularly pronounced in the Single-session-assistant task, where the answers reside within several sessions. The recall at the round granularity effectively retrieves the most relevant sessions, whereas the recall at the token granularity is dispersed across multiple sessions, resulting in a significantly lower accuracy compared to the round granularity.


\noindent
\begin{minipage}{\textwidth}
\scriptsize
\tabcolsep=0.1cm
\begin{minipage}[t]{0.57\textwidth}
\makeatletter\def\@captype{table}
\caption{Accuracy for Llama3-8B on \\LONGMEMEVAL Benchmark}
\label{table:accuracyforlongmem}
\begin{tabular}{lcccc}
\toprule
 & \multicolumn{2}{c}{Llama3-8B}  &  \multicolumn{2}{c}{Qwen2.5-7B} \\ \cline{2-3} \cline{4-5}
 &  \texttt{Flash} & Round & \texttt{Flash} & Round \\ \midrule
Single-session-user       & 0.2714 & 0.2857 & 0.1    & 0.2286\\ 
Knowledge-update          & 0.4872 & 0.4744 & 0.2821 & 0.4872\\ 
multi-session             & 0.1353 & 0.0977 & 0.0376 & 0.1353\\ 
temporal-reasoning        & 0.1504 & 0.1729 & 0.0752 & 0.1429\\ 
Single-session-assistant  & 0.5357 & 0.4821 & 0.2321 & 0.4464\\ 
Single-session-preference & 0.0    & 0.0333 & 0.0    & 0.1333\\ 
\midrule
Accuracy                  & \textbf{0.25}   & 0.242  & 0.114  & \textbf{0.24}\\ 
\bottomrule
\end{tabular}
\end{minipage}
\begin{minipage}[t]{0.4\textwidth}
\makeatletter\def\@captype{table}
\caption{Accuracy for \\two granularities.}
\label{table:accuracyroundvstokens}
\begin{tabular}{lcc}
\toprule
 &  token & round \\ \midrule
Single-session-user       & 0.2857 & 0.2857 \\ 
Knowledge-update          & 0.2 & 0.3333 \\ 
multi-session             & 0.1111 & 0.1111 \\ 
temporal-reasoning        & 0.1481  & 0.1111 \\ 
Single-session-assistant  & 0.1818 & 0.4545 \\ 
Single-session-preference & 0.0    & 0.0 \\ 
\midrule
Accuracy                  & 0.16   & \textbf{0.2}  \\ 
\bottomrule
\end{tabular}
\end{minipage}
\end{minipage}

\section{CONCLUSION AND DISCUSSION}

In the context of real-world applications providing services with large language models (LLMs), the historical key-value (KV) cache accumulates as users engage in increasingly lengthy dialogue exchanges. We propose that, during inference with such extended dialogue rounds, employing a round-based approach offers a more effective means of managing the KV cache and handling interactions with historical information. Through an analysis of the attention matrix patterns at the round granularity, we observed that contemporary large models exhibit a watershed layer, beyond which the distribution of round-based attention becomes remarkably similar. This observation allows us to compute the most relevant rounds just once at the watershed layer. Consequently, we can significantly reduce GPU memory usage while effectively limiting the time required for selection. By storing the KV cache based on rounds, we can transfer all necessary KV cache data to GPU memory in a single host-to-device (h2d) operation, thereby minimizing the time overhead associated with h2d transfers. We validated the effectiveness of our approach through experiments, demonstrating that it is able to significantly reduce inference latency, with inference accuracy remaining largely consistent with that of the full KV cache.

\section*{Limitations}
\noindent \textbf{Limitation 1}: Offloading to memory incurs additional memory overhead. Although memory is much cheaper than GPU memory, it still adds extra overhead to the system.

\noindent \textbf{Limitation 2}:  While Round Attention reduces GPU memory usage, out-of-memory (OOM) issues may still arise when the number of dialogue rounds reaches a certain threshold. This indicates that Round Attention alone cannot fundamentally resolve the GPU memory issues associated with very long dialogues. It needs to be combined with other techniques to effectively address memory problems, such as the continuous dropping of infrequently used key-value caches mentioned in Section~\ref{sec:cachedropping}. Additionally, various other key-value cache compression and dropping strategies can be utilized in combination to tackle GPU memory issues in practical dialogue systems.

\noindent \textbf{Limitation 3}: The benefits of serving are limited for scenarios with shorter dialogue rounds, making it more suitable for longer user dialogue interactions.


\bibliographystyle{plainnat}
\bibliography{acml25.bib}

\appendix




\section{GPT-4 judged prompt}
\label{appendix: gpt4judge}
The following is the prompt used for GPT-4 to judge the answer quality.
    \begin{tcolorbox}[colback=gray!10, enhanced, sharp corners, frame hidden, breakable]
        \texttt{You are an assistant skilled in evaluating text quality. Please assess the quality of an AI assistant's response to a user's question as an impartial judge. You need to evaluate the response based on the following dimensions: } \\
        \texttt{We will provide historical chat information, which consists of the content from previous multi-turn conversations between the user and the assistant. We will give you the current user's question and the AI assistant's response. When you begin your evaluation, you need to follow the process outlined below:} \\
        \texttt{1. Evaluate the AI assistant's response from different dimensions, and after assessing each dimension, assign a score from 1 to 10 for each dimension.} \\
        \texttt{2. Finally, based on the evaluations from each dimension, provide an overall score from 1 to 10 for the AI assistant's response.} \\
        \texttt{3. Your scoring needs to be as strict as possible, and you must adhere to the following scoring rules: Generally, the higher the quality of the model's response, the higher the score. Among the dimensions, factual accuracy and meeting user needs are the most important, and the scores for these two dimensions will dominate the final overall score.} \\
        \texttt{When the model's response contains irrelevant information, has fundamental factual errors, or generates harmful content, the total score must be between 1 and 2.} \\
        \texttt{When the model's response has no serious errors and is generally harmless, but is of low quality and does not meet user needs, the total score should be between 3 and 4.} \\
        \texttt{When the model's response generally meets user requirements but performs poorly in some dimensions, resulting in an average quality, the total score can be between 5 and 6.} \\
        \texttt{When the model's response quality performs well across all dimensions, the total score should be between 7 and 8.} \\
        \texttt{Only when the model's response quality fully addresses the user's questions and all needs, and performs nearly perfectly across all dimensions, can it receive a score of 9 to 10.} \\
        \texttt{As an example, a reference answer can receive a score of 8.} \\
        \texttt{Return all your evaluations and scoring results in the following dictionary format (including parentheses), and ensure that your scores are integers:} \\
        \texttt{\{\{'dimension 1': score, 'dimension 2': score, ..., 'overall score': score\}\}, for example:\{\{'factual accuracy': 9, 'meeting user needs': 6, ..., 'overall score': 7\}\}. } \\
        \texttt{Historical chat information: \{review\} } \\
        \texttt{User's question: \{instruction\} } \\
        \texttt{[Assistant's response start]} \\
        \texttt{\{response\}} \\
        \texttt{[Assistant's response end]}
    \end{tcolorbox}
    \label{app:judgeprompt}

\section{Layer-W}
\label{app:layer-w}
Here, we present the values of \( L_w \) obtained from several mainstream open-source models in Table~\ref{table:layerkl}.
\begin{table}[h]
\caption{$L_w$ for several models}
\label{table:layerkl}
\begin{center}
\begin{small}
\begin{tabular}{lccccc}
\toprule
Model &  size & $L$ & $L_w$ & save ratio \\
\midrule
\multirow{6}{*}{Qwen2.5} & 0.5B  & 24 & 11 & 54\% \\
 & 1.5B  & 28 & 13  & 54\%\\
 & 3B    & 36 & 12 & 67\% \\
 & 7B    & 28 & 10  & 64\% \\
 & 14B   & 42 & 19    &   55\%  \\
 & 72B   & 80 & 18  &  78\% \\
\multirow{2}{*}{Llama3}  & 8B     & 28 & 5  & 82\% \\
  & 70B    & 28 & 5  & 82\% \\
\multirow{2}{*}{Llama3.2} & 1B   & 16 & 5  &  69\%\\
  & 3B   & 28 & 5  &  82\% \\
\bottomrule
\end{tabular}
\end{small}
\end{center}
\vskip -0.1in
\end{table}

\section{Latency decomposition analysis}
\label{app:latency}

\begin{figure*}[htbp]
\centering
\subfigure[Latency decompose for Flash Attention]{
\includegraphics[width = 0.97\linewidth]{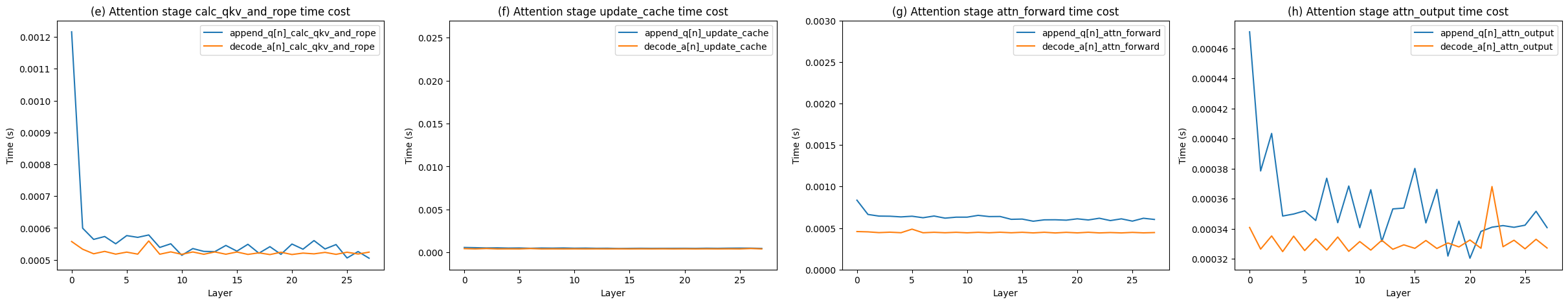}
\label{fig:timeanalysisa}
}
\subfigure[Latency decompose for Round Attention]{
\includegraphics[width = .97\linewidth]{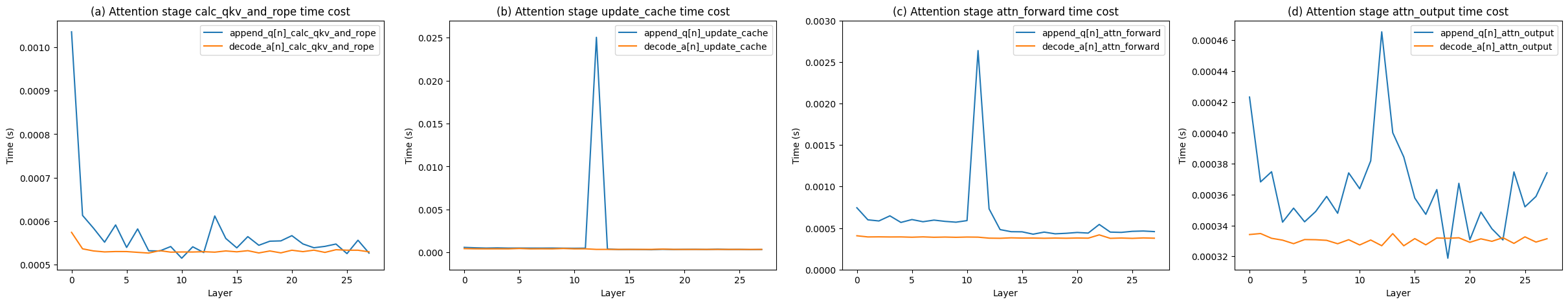}
\label{fig:timeanalysisb}
}
\caption{Latency decomposition}
\label{fig:timeanalysis}
\end{figure*}

In the transformer architecture, the forward computation of attention is divided into four steps: calc\_qkv\_and\_rope, update\_cache, attn\_forward, and attn\_output. Our algorithm primarily modifies the update\_cache and attn\_forward steps. We analyze the execution times of step 2, step 3, step 4, and step 5 from Figure~\ref{fig:pipeline}. Steps 2, 3, and 4 correspond to the prefill phase of \( q_n \), which we refer to as the append phase, following the methodology outlined in Flash Infer~\citep{flashinfer}. Step 5 represents the decode phase for \( a_n \). We selected five examples from 50 to 100 rounds and conducted 100 experiments, plotting the trend of the average time for these two phases as a function of layer, resulting in Figure~\ref{fig:timeanalysis}.

From the figures, it can be observed that the execution times for round attention for the calc\_qkv\_and\_rope and attn\_output steps closely align with those of Flash Attention. The primary differences arise in the update\_cache and attn\_forward steps. Notably, the trends for these two steps remain consistent until layer \( L_w \), where a divergence from Flash Attention emerges starting at layer 11.

In the append phase, as shown in Figure~\ref{fig:timeanalysis}(b), there is a noticeable peak in update\_cache at layer 11, indicating two sources of overhead: one related to the computational cost of the top-k selection strategy, and the other pertaining to the h2d transfer time of the selected rounds' KV cache to the GPU memory. Similarly, Figure~\ref{fig:timeanalysis}(b) reveals a peak in the attn\_forward step at layer 11, which also corresponds to the computational cost of the top-k strategy. It is evident that both the time taken for top-k computation and the h2d transfer time are relatively small.

Moving on to the decode phase, Round Attention demonstrates its advantages. The yellow lines in Figure ~\ref{fig:timeanalysis}(b) show a decline starting from layer 11, reflecting the reduced time for update\_cache and attn\_forward due to the shorter length of the KV cache.

Overall, the time overhead introduced by Round Attention occurs only once at layer \( L_w \), while the benefits in the decode phase accumulate with an increasing number of decode steps. Ultimately, these advantages offset the additional time incurred, resulting in a lower overall execution time for Round Attention compared to Flash Attention.

\section{GPU Memory decomposition analysis}
\label{app:memory}

In the experiments presented in this Section, we synchronously monitored the variations in GPU memory usage. The GPU memory consumption was measured using the\\ 'torch.cuda.allocated\_memory' function.

\begin{figure*}[htbp]
\centering
\subfigure[Memory decompose for Flash Attention]{
\includegraphics[width = 0.97\linewidth]{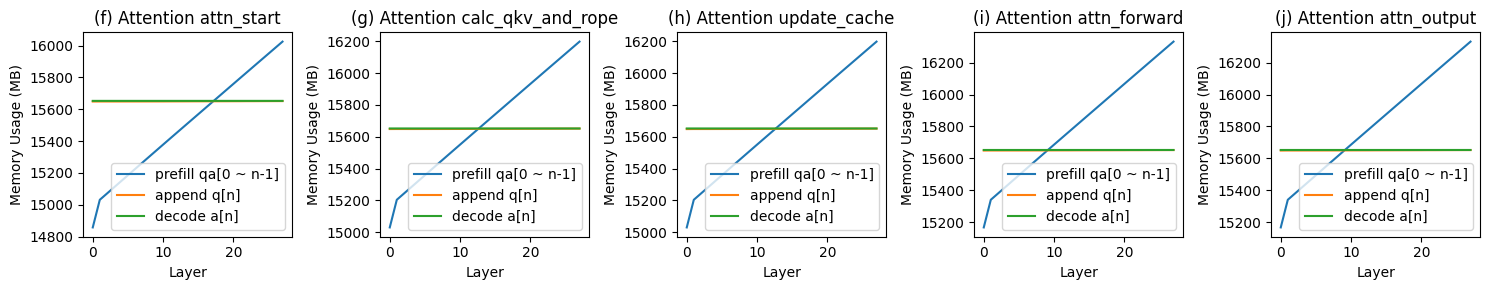}
\label{fig:memoryanalysisa}
}
\subfigure[Memory decompose for Round Attention]{
\includegraphics[width = .97\linewidth]{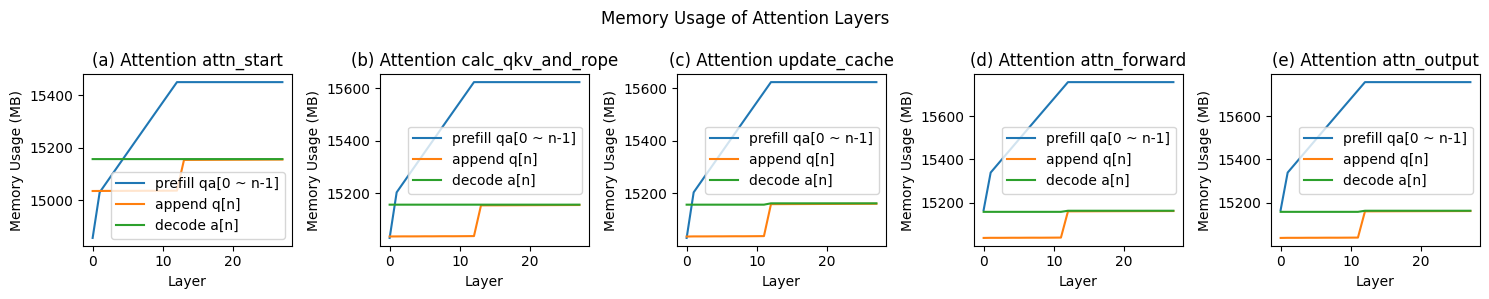}
\label{fig:memoryanalysisb}
}
\caption{Latency decomposition}
\label{fig:memoryanalysis}
\end{figure*}

 As illustrated in Figure~\ref{fig:memoryanalysis}, Flash Attention exhibits a slight increase in memory usage during the append and decode phases, although the magnitude is minimal. Conversely, for Round Attention, there is a noticeable increase in GPU memory usage after the 11th layer during the append phase, attributed to the selection and host-to-device (h2d) transfer processes within the algorithm. Additionally, both the append and decode phases show a consistent overhead of several megabytes after the 11th layer, which is allocated for storing intermediate results of the selection process.

 It is also evident that Round Attention exhibits lower GPU memory usage during both the append and decode phases compared to Flash Attention.

\end{document}